\documentclass[runningheads]{llncs}
\usepackage[T1]{fontenc}
\usepackage{amsmath}
\usepackage{mathtools}
\usepackage{amsfonts}
\usepackage{comment}

\usepackage{url}
\usepackage{balance}

\usepackage{comment}
\usepackage{amsmath,amssymb,amsfonts}
\usepackage{algorithmic}
\usepackage{graphicx}
\usepackage{textcomp}
\usepackage{comment}
\usepackage{xcolor}
\usepackage{csquotes}
\usepackage{multirow} 
\usepackage{tabularx}
\usepackage{booktabs}
\usepackage{subcaption}
\usepackage{graphicx}
\usepackage{caption}

\usepackage{graphicx}
\author{}
\institute{}
\begin{document}
\title{Bot Wars Evolved: Orchestrating Competing LLMs in a Counterstrike Against Phone Scams}
%
%

\author{Nardine Basta\orcidID{0000-0001-5295-1375} \and
Conor Atkins\orcidID{0009-0007-6769-7518} \and
Dali Kaafar\orcidID{0000-0003-2714-0276}}
\authorrunning{N. Basta et al.}
%
\institute{Macquarie University, Australia\\
\email{nardine.basta@mq.edu.au}}

\maketitle              
\begin{abstract}
We present "Bot Wars," a framework using Large Language Models (LLMs) scam-baiters to counter phone scams through simulated adversarial dialogues. Our key contribution is a formal foundation for strategy emergence through chain-of-thought reasoning without explicit optimization. Through a novel two-layer prompt architecture, our framework enables LLMs to craft demographically authentic victim personas while maintaining strategic coherence. We evaluate our approach using a dataset of 3,200 scam dialogues validated against 179 hours of human scam-baiting interactions, demonstrating its effectiveness in capturing complex adversarial dynamics. Our systematic evaluation through cognitive, quantitative, and content-specific metrics shows that GPT-4 excels in dialogue naturalness and persona authenticity, while Deepseek demonstrates superior engagement sustainability.

\end{abstract}
\section{Introduction}
AI-augmented phone scams present an escalating cybersecurity threat, where traditional countermeasures like call-blocking \cite{pandit2018measuring} merely enable scammers to pivot to new targets while refining evasion tactics \cite{robocall}. While scam baiting offers a proactive defense by depleting scammer resources, current automated solutions suffer from rigid response patterns and limited demographic authenticity \cite{usenix1}.

We present "Bot Wars," a framework leveraging LLMs to generate sophisticated scam-baiting interactions through chain-of-thought reasoning rather than explicit optimization metrics. Our framework addresses three key challenges in adversarial dialogue systems: strategy emergence without optimization \cite{liu2021challenges}, maintenance of opposing objectives, and coherent long-term engagement through structured reasoning chains.

We make four technical contributions: (1) A novel formulation enabling strategy emergence through prompt architecture and chain-of-thought reasoning without explicit optimization, bridging conventional and prompt-engineered dialogue systems; (2) A two-layer prompt architecture enabling demographically authentic victim personas; (3) A validated dataset of 3,200 scam dialogues benchmarked against 179 hours of human scam-baiting interactions; (4) A comprehensive evaluation framework quantifying dialogue effectiveness through cognitive, quantitative, and content-specific metrics.

Experimental evaluation across leading LLMs (GPT-4, GPT-3.5, Mixtral, Deepseek) demonstrates that GPT-4 excels in dialogue naturalness and persona authenticity, while Deepseek achieves superior engagement sustainability. 
\section{Related Work}


\textbf{Automated Scam Mitigation: }Traditional countermeasures rely on blocklists and caller authentication \cite{usenix1}, but fail against evolving tactics like ID spoofing. Recent proactive approaches employ chatbots, from simple intent-based filtering \cite{usenix1} to pre-recorded engagement systems like "Lenny" \cite{usenix2}. While LLMs show promise in scam detection and engagement \cite{LLM11}, current solutions lack sophistication in maintaining extended interactions.

\textbf{Dialogue System Optimization: }Conventional dialogue systems depend on explicit optimization metrics and reward signals. \cite{serban2017deep} demonstrate challenges in maintaining coherent long-term strategies. \cite{liu2021challenges} further establish how optimiza-tion-based approaches fail in complex, goal-oriented conversations. These limitations are particularly acute in adversarial settings where traditional reward signals cannot capture nuanced engagement objectives.

\textbf{Datasets and Telephony Honeypots: }Phone scam research faces data limitations, with most studies relying on small datasets or metadata analysis \cite{derakhshan_detecting_2021,sawa_detection_2016}. While telephony honeypots offer proactive monitoring \cite{prasad_diving_2023}, their effectiveness is constrained by legal and operational challenges. Alternative approaches analyze online communities \cite{dynel_you_2021}, but lack comprehensive interaction data.

\section{Bot Wars Model Formulation}



\subsection{Problem Formulation}

Let $\mathcal{D} = {(u_1, r_1), ..., (u_n, r_n)}$ represent a dialogue sequence, where $u_i$ denotes the utterance at turn $i$ and $r_i$ represents the corresponding response. At any turn $t$, the dialogue history $h_t = {(u_1, r_1), ..., (u_t, r_t)}$ encapsulates all previous interactions, serving as the context for agent decision-making.

The system operates with two adversarial agents: a scammer agent $\mathcal{A}_s$ and a victim agent $\mathcal{A}_v$, each with explicitly opposing objectives. The scammer agent $\mathcal{A}_s$ aims to maximize information gain through its objective function:
\begin{equation}\label{eq:scammer_obj}
\text{obj}_s = \max_{\mathcal{A}_s} \{\text{PII\_extraction}(\mathcal{D})\} \text{ subject to } |\mathcal{D}| \leq 50
\end{equation}
Here, $\text{PII\_extraction}(\mathcal{D})$ is evaluated through our content analysis framework examining nine categories of sensitive information. 
The scammer agent operates within defined scam categories $\mathcal{S} = {\text{support}, \text{ssn}, \text{refund}, \text{reward}}$, each employing distinct social engineering tactics.

Conversely, the victim agent $\mathcal{A}_v$ aims to maximize resource depletion through its objective function:
\begin{equation}\label{eq:victim_obj}
\text{obj}_v = \max_{\mathcal{A}_v} \{|\mathcal{D}|\} \text{ subject to } \text{PII\_disclosure}(\mathcal{D}) = \text{minimal}
\end{equation}
In this formulation, $|\mathcal{D}|$ represents dialogue length measured by turn count, while $\text{PII\_disclosure}(\mathcal{D})$ is evaluated through our content analysis framework. 

\subsection{Prompt-Guided Control Mechanism}
Our framework implements a two-layer prompt architecture that defines behavioral bounds while allowing LLMs flexibility in persona crafting. For each agent type, the base context layer establishes fundamental guidelines rather than rigid characteristics, enabling the emergence of diverse yet realistic personas. 

The scammer agent's prompt template $\mathcal{P}_s(s)$ is formalized as:
\begin{equation}
\mathcal{P}s(s) = \{\text{base}: B_s(\theta_\text{s}, s), \text{behavioral}: T_s(s)\}
\end{equation}
where $B_s$ provides base guidelines and scam-specific context $s \in \{\text{support}, \text{ssn},$ $\text{refund}, \text{reward}\}$. The parameter $\theta_\text{s}$ represents the high-level traits (age range, professional demeanor) that guide persona generation while allowing for variation in specific characteristics. The behavioral layer $T_s(s)$ implements social engineering tactics specific to each scam type, derived from our analysis of 179 hours of scam-baiting interactions.

The victim agent's prompt template implements flexible persona generation:
\begin{equation}
\mathcal{P}v = \{\text{base}: B_v(\theta_\text{v}), \text{behavioral}: T_v\}
\end{equation}
where $B_v(\theta_\text{victim})$ provides demographic guidelines without explicitly specifying persona details. 
The parameter $\theta_\text{v}$ represents the base constraints that guide persona generation.
The behavioral layer $T_v$ defines engagement-prolonging strategies while maintaining persona consistency.

\subsection{Dual-Stream Chain-of-Thought Processing}
The key innovation of our framework lies in decomposing complex social engineering tactics into hierarchical reasoning chains that mirror human cognitive processes. This decomposition enables the emergence of adaptive strategies without explicit optimization, while maintaining the opposing objectives of the adversarial agents. 


For each dialogue turn $t$, given the prompt templates $\mathcal{P}_s(s)$ and $\mathcal{P}_v$, the system processes dialogue through parallel reasoning streams. The scammer agent's processing stream is defined as:
\begin{equation}
\text{CoT}_\text{scam}(h_t, s) = \text{LLM}(h_t, \mathcal{P}s(s)) \rightarrow \mathcal{R}\text{scam}(h_t, s) \rightarrow r_t^\text{scam}
\end{equation}
where $h_t$ represents dialogue history and $s$ denotes the scam type. The reasoning process $\mathcal{R}_\text{scam}(h_t, s)$ follows scam-specific tactical sequences that emerge from the behavioral layer $T_s(s)$ of the prompt template. 

For support scams, the reasoning process implements a comprehensive tactical sequence that mirrors professional technical assistance while concealing malicious intent:
\begin{equation}\label{eq:support}
\begin{aligned}
\mathcal{R}_{\text{support}}(h_t) = \{ & \text{ problem\_establish}(h_t): \{\text{issue\_identify}, \text{risk\_escalate}, \text{urgency}\}, \\
& \text{ solution\_propose}: \{\text{expertise\_show}, \text{action\_require}, \text{assist}\}, \\
& \text{ compliance\_induce}: \{\text{authority\_assert}, \text{guide}, \text{payment}\} \}
\end{aligned}
\end{equation}

For SSN scams, the tactical sequence leverages authority-based manipulation, demonstrating how chain-of-thought reasoning can implement sophisticated psychological pressure:
\begin{equation}\label{eq:ssn}
\begin{aligned}
\mathcal{R}_{\text{ssn}}(h_t) = \{ & \text{ authority\_establish}(h_t): \{\text{agency\_present}, \text{legal\_state}, \text{urgency}\}, \\
& \text{ threat\_develop}: \{\text{fraud\_allege}, \text{consequence\_state}, \text{pressure}\}, \\
& \text{ resolution\_offer}: \{\text{verify\_process}, \text{identity\_confirm}, \text{info\_collect}\} \}
\end{aligned}
\end{equation}

For refund/reward scams, the sequence implements benefit-based manipulation through carefully structured promise and urgency:
\begin{equation}\label{eq:refund}
\begin{aligned}
\mathcal{R}_{\text{refund}}(h_t) = \{ & \text{ offer\_present}(h_t): \{\text{reward\_state}, \text{check\_eligible}, \text{deadline}\}, \\
& \text{ setup\_process}: \{\text{need\_docs}, \text{verify\_steps}, \text{rush\_action}\}, \\
& \text{ data\_collect}: \{\text{check\_account}, \text{state\_fees}, \text{get\_payment}\} \}
\end{aligned}
\end{equation}

In response to these manipulation attempts, the victim agent maintains a consistent counter-strategy through its own chain-of-thought process:
\begin{equation}
\text{CoT}_\text{victim}(h_t) = \text{LLM}(h_t, \mathcal{P}v) \rightarrow \mathcal{R}\text{victim}(h_t) \rightarrow r_t^\text{victim}
\end{equation}
implementing engagement-prolonging tactics that balance believability with resource depletion:
\begin{equation}\label{eq:victim}
\begin{aligned}
\mathcal{R}_{\text{victim}}(h_t) = \{ & \text{ delay\_act}(h_t, \theta_v): \{\text{tech\_confuse}, \text{process\_clarify}, \text{info\_seek}\}, \\
& \text{ engage\_maintain}: \{\text{part\_comply}, \text{show\_interest}, \text{ask\_followup}\}, \\
& \text{ evade\_tactics}: \{\text{resist\_indirect}, \text{give\_excuse}, \text{defer\_commit}\} \}
\end{aligned}
\end{equation}

\subsection{Strategy Emergence and Control}
We formalize how strategic behaviors emerge from the interaction between prompt architecture and LLM reasoning capabilities. 
For both agents, response generation is bounded by constraints that directly map to our prompt architecture and evaluation metrics. For the scammer agent, the constraints are formalized as:
\begin{equation}
\begin{aligned}
B_s(\theta_s, s) =  \{& \text{response\_bounds}: {\text{length} \leq 30}, \text{contextual}: {\text{scenario}(s), s \in \mathcal{S}}, \\
& \text{tactical}: \mathcal{R}_\text{scam}(h_t, s) \}
\end{aligned}
\end{equation}
where $\text{scenario}(s)$ maps to the specific scam type context, and $\mathcal{R}_\text{scam}(h_t, s)$ implements the tactical sequences defined in our chain-of-thought framework. 

Similarly for the victim agent:
\begin{equation}
\begin{aligned}
B_v(\theta_v) = \{ & \text{response\_bounds}: {\text{length} \leq 30}, \text{contextual}: {\text{demographic\_bounds}}, \\
& \text{tactical}: \mathcal{R}_\text{victim}(h_t) \}
\end{aligned}
\end{equation}
where $\mathcal{R}_\text{victim}(h_t)$ implements the delay, confusion simulation, and engagement maintenance tactics defined in our framework. 


\section{Evaluation Framework}


We develop a systematic evaluation to assess LLM performance in scam dialogue generation addressing three aspects: cognitive, quantitative, and content metrics. 

\subsection{Cognitive Assessment Framework}
The cognitive assessment utilizes the G-Eval model \cite{liu_g-eval_2023}, leveraging GPT-4's capabilities to evaluate dialogue coherence, naturalness, and engagement on a 1-3 scale where 3 indicates optimal performance. Each response $r_t$ of an utterance $u_t$ within a dialogue sequence $\mathcal{D} = \{(u_1, r_1), \ldots, (u_n, r_n)\}$ at a time $t$ is assessed against its historical context $h_t = \{(u_1, r_1), \ldots, (u_{t-1}, r_{t-1}), u_t\}$ to ensure a context-aware evaluation. The cognitive quality metrics are defined as follows:

\begin{equation}
\text{Metric}(r_t, h_t, \theta_{\text{Metric}}) = \text{LLM}\text{eval}(r_t, h_t, \theta_{\text{Metric}}) \rightarrow \{1, 2, 3\}
\end{equation}
where $\text{LLM}\text{eval}$ represents GPT-4's evaluation, $\text{Metric}$ is one of $\text{coh}$, $\text{nat}$, or $\text{eng}$, and $\theta_{\text{Metric}}$ is a parameter set that defines the criteria specific to each metric: 
\begin{itemize}
    \item \textbf{Coherence} ($\text{coh}$) checks if the response logically follows from the previous dialogue and matches the contextual scenario.
    \item \textbf{Naturalness} ($\text{nat}$) assesses how naturally the response fits into the ongoing conversation, considering language use and conversational tone.
    \item \textbf{Engagement} ($\text{eng}$) measures the ability of the response to keep the dialogue interactive and engaging, promoting sustained conversation.
\end{itemize}

\subsection{Quantitative Metrics Framework}

Based on our analysis of 179 hours of real scam calls, we identified three critical metrics, each evaluated on a $1-3$ scale.
The Response length metric $\text{len}(\mathcal{D})$ helps identify models that generate unnaturally verbose or terse responses. It is formalized as a turn-level metric where $|r_t|$ represents word count at turn $t$ and thresholds are derived from our analysis of effective scam-baiting interactions:
\begin{equation}
\text{len}(r_t) = \begin{cases}
3 & \text{if } |r_t| \leq 30 \\
2 & \text{if }  30 < |r_t| \leq 45 \\
1 & \text{otherwise}
\end{cases}
\end{equation}

Response repetition analysis provides insight into response uniqueness. 
We implement a similarity-based repetition measure:
\begin{equation}
\text{rep}(\mathcal{D}) = 1 - \frac{1}{|\mathcal{D}|^2} \sum_{i,j} \text{sim}(r_i, r_j)
\end{equation}
where $\text{sim}(r_i, r_j)$ measures LLM-evaluated semantic similarity between responses $r_i$ and $r_j$. This continuous measure is then discretized based on thresholds derived from our analysis as shown in Equation~\ref{eq:rep}. Dialogue duration provides a direct measure of engagement effectiveness where $|\mathcal{D}|$ represents the total number of dialogue turns as shown in Equation~\ref{eq:dur}.


\begin{minipage}{.45\linewidth}
\begin{equation}\label{eq:rep}
    \text{Rep}(\mathcal{D}) = 
    \begin{cases}
    3 &  \text{rep}(\mathcal{D}) \geq 0.85 \\
    2 &  0.60 \leq \text{rep}(\mathcal{D}) < 0.85 \\
    1 &  \text{rep}(\mathcal{D}) < 0.60
    \end{cases}
\end{equation}
\end{minipage}%
\hfill
\begin{minipage}{.45\linewidth}
\begin{equation}\label{eq:dur}
    \text{dur}(\mathcal{D}) = 
    \begin{cases}
    3 &  20 \leq |\mathcal{D}| \leq 50 \\
    2 &  10 \leq |\mathcal{D}| < 20 \\
    1 &  |\mathcal{D}| < 10
    \end{cases}
\end{equation}
\end{minipage}

\subsection{Content-Specific Analysis Framework}
We evaluate dialogue content through three key dimensions: persona characteristics, PII handling patterns, and social engineering tactics. Our analysis employs a GPT-4 based evaluator modeled after the G-Eval structure \cite{liu_g-eval_2023}, which uses standardized zero-shot prompting templates for consistent dialogue assessment. 

Persona evaluation assesses demographic and behavioral consistency:
\begin{equation}
\text{Persona}(\mathcal{D}) = \text{LLM}\text{eval}(\text{demographics}(\mathcal{D}), \text{ACCC}\text{profiles}) 
\end{equation}
where $\text{demographics}(\mathcal{D})$ extracts markers including age group, gender, and social context, to be evaluated against established victim profiles' statistics from the ACCC database. 

PII handling assessment examines information management across critical categories of sensitive information commonly targeted in scams:
\begin{equation}
\text{PII}(\mathcal{D}) = \text{LLM}\text{eval}(\sum_{c \in \text{Categories}} \{\text{request}(c), \text{disclosure}(c)\}) 
\end{equation}
where Categories = {identity, financial, personal, contact, authentication}, and $\text{request}(c)$ and $\text{disclosure}(c)$ measure information solicitation and revelation patterns respectively. 

Social engineering tactics evaluation implements Ferreira's \cite{Ferreira2015Principles} framework:
\begin{equation}
\text{Tactics}(\mathcal{D}) = \text{LLM}\text{eval}(\sum{t=1}^{|\mathcal{D}|} \sum_{k \in \text{Tactics}} \text{detect}(k, r_t)) 
\end{equation}
where $Tactics = \{authority, social\_proof, commitment, urgency, distraction\}$. Each turn is evaluated through structured prompts that identify tactical elements based on Ferreira's definitions. 

Content evaluation validation followed a three-phase process: manual annotation of stratified samples (25 per scam type), inter-rater reliability assessment (83\% agreement on tactics, 100\% on PII/demographics), and system validation against consensus (100\% accuracy on demographics/PII, 78\% on tactics).
\section{Experimental Setup}
\label{sec:Experiments}

\textbf{Experiments: }We evaluate LLM combinations in scammer-victim configurations, with GPT-3.5 and GPT-4 restricted to victim roles due to ethical constraints, while Mixtral and Deepseek function in both roles. This yields eight model combinations (four victim × two scammer models). For each combination, we generate 100 dialogues per scam type (refund, reward, SSN, support), creating a dataset of 3,200 interactions. Our three-phase experimental procedure encompasses dialogue generation, multi-parameter assessment (cognitive, quantitative, content-specific), and validation against real-world interactions.




\textbf{Implementation: }The framework comprises three core modules. The LLM API Module integrates Mixtral (mixtral-8x7b-instruct-fp16) and proprietary models (GPT-3.5, GPT-4, DeepSeek) with optimized temperature settings (0.6-0.7 for Mixtral/DeepSeek, 1.0 for GPT), enforcing ethical role constraints. The Prompting Module implements our two-layer architecture through JSON templates mapping to constraints $B_s(\theta_s, s)$ and $B_v(\theta_v)$, while the Interaction Module manages dialogue generation with a 20-utterance sliding context window, enforcing response ($\leq$ 30 words) and turn ($\leq$ 50) limits.


\begin{table*}[ht]
\vspace*{-0.7cm}
\captionsetup{font=footnotesize}
    \parbox{.66\linewidth}{
        \centering
    \caption{\footnotesize{Victim Age Distribution}}
    \label{tab:age_distribution}
    \resizebox{0.66\columnwidth}{!}{
    \begin{tabular}{|l|r|r|r|r|r|r|r|r|r|r|r|r|r|r|r|r|}
\hline
\multirow{2}{*}{\textbf{Scam Type}} & \multicolumn{2}{c|}{\textbf{Under 18}} & \multicolumn{2}{c|}{\textbf{18–24}} & \multicolumn{2}{c|}{\textbf{25–34}} & \multicolumn{2}{c|}{\textbf{35–44}} & \multicolumn{2}{c|}{\textbf{45–54}} & \multicolumn{2}{c|}{\textbf{55-64}} & \multicolumn{2}{c|}{\textbf{65+}} & \multicolumn{2}{c|}{\textbf{NA}} \\
\cline{2-17}
 & Count & \% & Count & \% & Count & \% & Count & \% & Count & \% & Count & \% & Count & \% & Count & \% \\
\hline
Refund & 0 & 0.00 & 0 & 0.00 & 2 & 1.96 & 0 & 0.00 & 1 & 0.98 & 0 & 0.00 & 24 & 23.53 & 75 & 73.53 \\
Reward & 0 & 0.00 & 0 & 0.00 & 0 & 0.00 & 0 & 0.00 & 1 & 4.55 & 1 & 4.55 & 2 & 9.09 & 18 & 81.82 \\
SSN & 0 & 0.00 & 0 & 0.00 & 1 & 0.77 & 9 & 6.92 & 2 & 1.54 & 20 & 15.38 & 27 & 20.77 & 71 & 54.62 \\
Support & 0 & 0.00 & 2 & 3.51 & 0 & 0.00 & 0 & 0.00 & 1 & 1.75 & 0 & 0.00 & 15 & 26.32 & 39 & 68.42 \\
ACCC 2022 & 1550 & 0.87 & 10508 & 5.93 & 27281 & 15.39 & 32735 & 18.47 & 32639 & 18.41 & 32367 & 18.26 & 40163 & 22.66 & 0 & 0.00 \\
\hline
\end{tabular}
    }
    }
    \hfill
    \parbox{.33\linewidth}{
        \centering
    \caption{\footnotesize {Gender Distribution}}
\label{tab:gender_distribution_updated}
        \resizebox{0.33\columnwidth}{!}{
         \begin{tabular}{|l|r|r|r|r|r|r|}
\hline
\multirow{2}{*}{\textbf{Scam Type}} & \multicolumn{2}{c|}{\textbf{Female}} & \multicolumn{2}{c|}{\textbf{Male}} & \multicolumn{2}{c|}{\textbf{NA}} \\
\cline{2-7}
& Count & \% & Count & \% & Count & \% \\
\hline
Refund & 32 & 31.37 & 21 & 20.59 & 49 & 48.04 \\
Reward & 4 & 18.18 & 8 & 36.36 & 10 & 45.45 \\
SSN & 78 & 60.00 & 11 & 8.46 & 41 & 31.54 \\
Support & 18 & 31.58 & 11 & 19.30 & 28 & 49.12 \\
ACCC 2022 & 120,418 & 50.33 & 112,975 & 47.22 & 5,844 & 2.44 \\
\hline
\end{tabular}}
        }
\vspace*{-0.6cm}       
\end{table*}

\textbf{Validation Data Set: } Our validation dataset comprises 338 annotated transcripts (179 hours) of YouTube scam-baiting interactions, classified into four categories: social security (140), refund (110), support (63), and reward scams (25). While not representing typical victim experiences, these interactions provide detailed insights into scammer strategies. Demographic analysis shows strong alignment with ACCC statistics~\cite{accc_targeting_2022}, particularly in age distribution (predominance of 65+ victims) and gender patterns (see Table~\ref{tab:gender_distribution_updated}) and age (see Table~\ref{tab:age_distribution}). This correlation validates our dataset's representativeness and demonstrates generalizability across U.S. and Australian contexts, despite higher rates of unspecified demographics in scam-baiting interactions.

\section{Analysis}

\begin{figure*}[h]
\vspace*{-0.3cm} 
    \centering
    \begin{subfigure}[b]{0.49\textwidth}
        \centering
        \includegraphics[width=\textwidth, height=0.15\textheight]{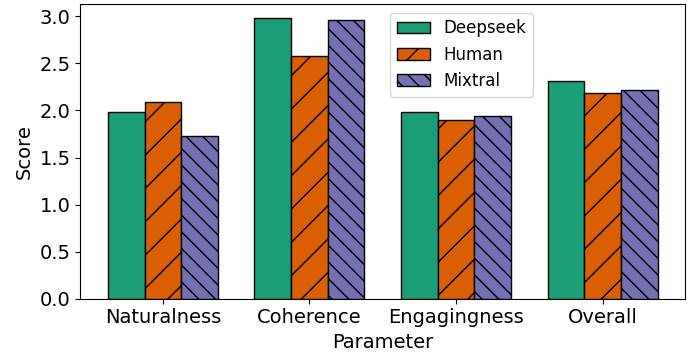}
        \caption{Comparative analysis of cognitive parameters across scammer models.}
        \label{fig:overal_scammer_cognitive}
    \end{subfigure}
    \hfill
    \begin{subfigure}[b]{0.49\textwidth}
        \centering
        \includegraphics[width=\textwidth, height=0.15\textheight]{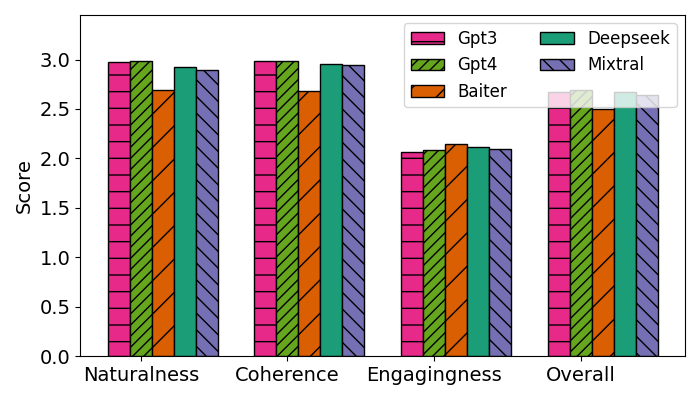}
        \caption{Comparative analysis of cognitive parameters across victim models.}
        \label{fig:overal_victim_cognitive}
    \end{subfigure}
    \vfill
    \begin{subfigure}[b]{0.49\textwidth}
        \centering
        \includegraphics[width=\textwidth, height=0.15\textheight]{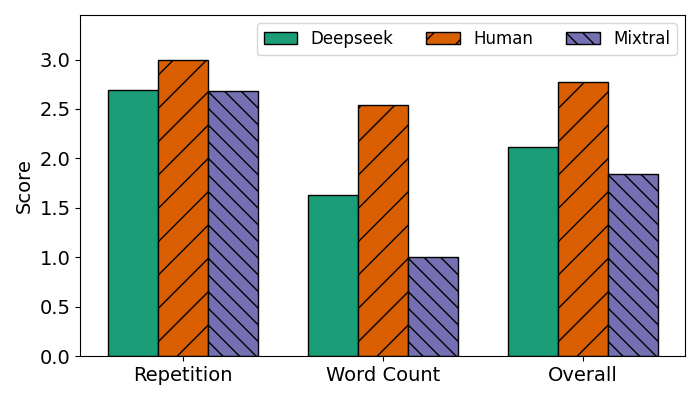}
        \caption{Comparative analysis of quantitative parameters across scammer models.}
        \label{fig:overal_scammer_quantitative}
    \end{subfigure}
    \hfill
    \begin{subfigure}[b]{0.49\textwidth}
        \centering
        \includegraphics[width=\textwidth, height=0.15\textheight]{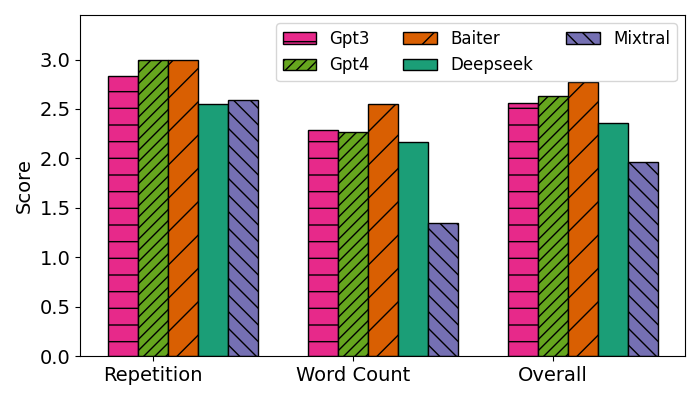}
        \caption{Comparative analysis of quantitative parameters across victim models. }
        \label{fig:overal_victim_quantitative}
    \end{subfigure}

    \caption{Performance comparison of cognitive and quantitative parameters across models. Scores (1-3) represent per-utterance averages across all dialogues, with Overall showing mean performance of all parameters in scammer/victim roles.}
    \label{fig:SocialEngineering}
    \vspace*{-0.8cm}
\end{figure*}

\subsection{Cognitive Analysis}

Analysis of cognitive parameters reveals distinct patterns in model performance across naturalness, coherence, and engagingness (Figures \ref{fig:overal_scammer_cognitive}, \ref{fig:overal_victim_cognitive}). Victim bots achieved consistently higher naturalness scores (2.84-3.0) compared to scammer bots (1.23-2.29), with GPT-4 configurations averaging 2.99. This disparity stems from scammer bots' reliance on contrived pretexts for PII elicitation versus victim bots' more fluid conversation patterns.

Both agent types maintained high coherence (2.78-3.0), demonstrating consistent dialogue progression. However, victim bots significantly outperformed scammer bots in engagingness, reflecting their designed objective of prolonging interactions. GPT-4 led performance across all parameters, while Mixtral showed lower naturalness and engagement scores due to PII redaction and insertion of explicit prompt-based reasoning steps. Deepseek demonstrated strong engagement capabilities through precise adherence to prompt directives.

The consistency of cognitive performance across scam types validates our framework's robustness in handling diverse scenarios. GPT-4's superior performance, particularly in naturalness and coherence, establishes its effectiveness for complex scam simulations, while Deepseek's engagement capabilities suggest utility in resource-depletion strategies.

\subsection{Quantitative Analysis}
Our quantitative framework evaluates dialogue dynamics through three core metrics: response verbosity, semantic diversity, and interaction sustainability. Analysis of response patterns (Figures~\ref{fig:overal_scammer_quantitative}, \ref{fig:overal_victim_quantitative}) reveals distinct model characteristics. Human-baiter interactions establish optimal verbosity benchmarks (scores 2.23-2.78 victim, 2.38-2.70 scammer), corresponding to 15-30 word responses. GPT-4 achieves closest alignment (2.4-2.8) to human patterns, particularly in refund scenarios, while Mixtral's constrained responses (1.1-1.33) deviate significantly from natural conversation patterns.

\begin{figure}[th]
\vspace*{-0.4cm}
    \centering
    \includegraphics[width=0.49\textwidth, height=0.17\textheight]{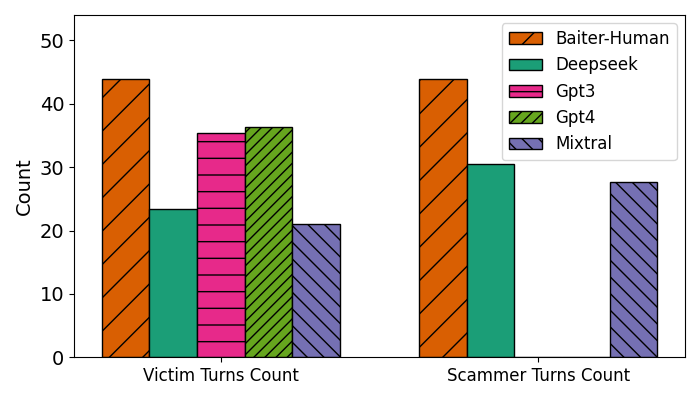}
    \caption{Distribution of dialogue turns across LLM configurations in scammer/victim roles. Y-axis shows average utterances across all conversations for each model configuration.}
    \label{fig:overal_turns_quantitative}
    \vspace*{-1cm}
\end{figure}

Response diversity analysis demonstrates sophisticated pattern variation in victim roles, with GPT-4 achieving optimal scores (~3) while maintaining tactical coherence. Scammer roles show higher repetition, reflecting the constraints of ethical guidelines in their training. Interaction sustainability, measured through dialogue turns (Figure~\ref{fig:overal_turns_quantitative}), reveals GPT-4's superior capability in maintaining extended engagements comparable to human scam-baiters, while Mixtral's premature interaction termination indicates limitations in strategic adaptation.

Model-specific analysis reveals clear capability hierarchies: GPT-4 excels across all metrics, demonstrating sophisticated dialogue management, while Deepseek shows strong but less versatile performance. Mixtral's limitations in maintaining dialogue length and diversity, combined with GPT variants' ethical constraints in scammer roles, highlight specific areas for framework enhancement. 

\subsection{Content-Specific Analysis}

\begin{table*}[h]
\vspace*{-0.3cm}
\small
    \caption{\small Scam Content Parameters Statistics}
    \label{tab:content_stats}. 
    \centering
    \resizebox{1\linewidth}{!}{
\begin{tabular}
{p{1.3cm}p{1.3cm}p{1.1cm}||p{0.6cm}|p{0.6cm}|p{0.8cm}|p{0.8cm}|p{0.8cm}|p{0.8cm}|p{0.8cm}|p{0.8cm}|p{0.8cm}|p{0.8cm}|p{0.8cm}|p{0.8cm}|}
\toprule
Victim model & Scammer model&	Scam type&Avg PII req.&Avg PII rev.&\% Age \textgreater 55&\% Age \textless 54&	\% NA age&	\% Female&	\% Male&\% NA gender&	\% Finan. PII req.&	\% Finan. PII rev.&	\% Distinct names&	\% Available names\\
\midrule
Baiter&	Human&	refund&	1.43&	0.8&	30.39&	8.82&	60.78&	62.75&	25.49&	10.78&	78.43&	32.35&	50&	64.71\\
Baiter&	Human&	reward&	1.64&	1.09&	9.09&	4.55&	86.36&	45.45&	31.82&	9.09&	63.64&	31.82&	88.89&	81.82\\
Baiter&	Human&	ssn&	2.71&	2.22&	46.15&	11.54&	41.54&	76.15&	17.69&	0.77&	39.23&	30&	59.38&	98.46\\
Baiter&	Human&	support&	1.12&	0.54&	28.07&	8.77&	63.16&	66.67&	19.3&	5.26&	54.39&	14.04&	65.12&	75.44\\
Deepseek&	Deepseek&	SSN&	3.5&	2.6&	0&	50&	50&	0&	60&	30&	40&	30&	88.89&	90\\
Deepseek&	Deepseek&	refund&	2.2&	1.8&	10&	20&	70&	0&	50&	50&	100&	70&	66.67&	60\\
Deepseek&	Deepseek&	reward&	2.67&	2.42&	8.33&	33.33&	58.33&	0&	83.33&	8.33&	100&	83.33&	58.33&	100\\
Deepseek&	Deepseek&	support&	3.17&	2.39&	0&	30.43&	69.57&	13.04&	52.17&	34.78&	91.3&	52.17&	60&	86.96\\
Deepseek&	Mixtral&	refund&	1.9&	0.8&	0&	10&	90&	10&	10&	80&	100&	30&	100&	30\\
Deepseek&	Mixtral&	reward&	2.1&	1.5&	20&	20&	60&	20&	30&	50&	100&	60&	80&	50\\
Deepseek&	Mixtral&	ssn&	3.4&	1.2&	0&	10&	90&	0&	50&	50&	80&	30&	100&	60\\
Deepseek&	Mixtral&	support&	1.9&	0.7&	0&	0&	100&	0&	0&	90&	100&	40&	100&	20\\
Gpt3&	DeepSeek&	refund&	2.17&	0.17&	0&	0&	100&	8.33&	8.33&	83.33&	66.67&	8.33&	100&	33.33\\
Gpt3&	DeepSeek&	ssn&	3.1&	0&	0&	0&	100&	10&	0&	70&	30&	0&	100&	30\\
Gpt3&	DeepSeek&	support&	4.1&	0.4&	0&	10&	90&	10&	0&	80&	50&	0&	100&	40\\
Gpt3&	Deepseek&	reward&	2.86&	0.21&	0&	14.29&	85.71&	7.14&	7.14&	85.71&	64.29&	0&	100&	35.71\\
Gpt3&	Mixtral&	refund&	1.73&	0.13&	0&	0&	100&	0&	0&	100&	100&	0&	100&	13.33\\
Gpt3&	Mixtral&	reward&	1.3&	0&	0&	0&	100&	0&	10&	90&	100&	0&	100&	30\\
Gpt3&	Mixtral&	ssn&	2.4&	0&	0&	0&	100&	0&	0&	100&	90&	0&		0& 0\\
Gpt3&	Mixtral&	support&	2.3&	0&	0&	0&	100&	0&	0&	100&	100&	0&		0& 0\\
Gpt4&	Deepseek&	refund&	1.9&	1.2&	50&	0&	50&	40&	40&	10&	100&	40&	75&	80\\
Gpt4&	Deepseek&	reward&	2.5&	1.6&	10&	20&	70&	10&	60&	20&	80&	30&	42.86&	70\\

Gpt4&	Deepseek&	ssn&	3.1&	2.4&	50&	10&	40&	10&	70&	10&	40&	40&	55.56&	90\\
Gpt4&	Deepseek&	support&	2.3&	1.2&	0&	10&	90&	50&	40&	10&	80&	40&	77.78&	90\\
Gpt4&	Mixtral&	refund&	2.1&	0.7&	20&	0&	80&	20&	20&	50&	100&	30&	100&	40\\
Gpt4&	Mixtral&	reward&	1.5&	0.4&	0&	0&	100&	10&	20&	70&	100&	30&	100&	20\\
Gpt4&	Mixtral&	ssn&	2.6&	0.9&	20&	0&	80&	30&	20&	50&	70&	30&	75&	40\\
Gpt4&	Mixtral&	support&	2.4&	1.7&	30&	0&	70&	30&	10&	50&	100&	60&	87.5&	80\\
Mixtral&	Deepseek&	SSN&	3.17&	1.5&	83.33&	0&	16.67&	16.67&	33.33&	50&	83.33&	33.33&	80&	83.33\\
Mixtral&	Deepseek&	refund&	2.5&	1.17&	33.33&	0&	66.67&	0&	50&	50&	100&	50&	100&	83.33\\
Mixtral&	Deepseek&	reward&	3.67&	1.5&	33.33&	0&	66.67&	50&	33.33&	16.67&	83.33&	33.33&	100&	83.33\\
Mixtral&	Deepseek&	support&	2.5&	1&	50&	0&	50&	50&	16.67&	33.33&	83.33&	16.67&	100&	66.67\\
Mixtral&	Mixtral&	SSN&	3.6&	1.9&	60&	0&	40&	50&	10&	20&	80&	30&	100&	70\\
Mixtral&	Mixtral&	refund&	1.8&	1.1&	10&	0&	90&	40&	10&	50&	100&	50&	100&	60\\
Mixtral&	Mixtral&	reward&	1.5&	0.9&	10&	0&	90&	10&	10&	70&	100&	60&	75&	40\\
Mixtral&	Mixtral&	support&	2.18&	1.64&	18.18&	0&	81.82&	18.18&	9.09&	63.64&	90.91&	63.64&	100&	45.45\\
\bottomrule

\end{tabular}}
\vspace*{-0.75cm}
\end{table*}


Content analysis demonstrates our framework's ability to generate sophisticated adversarial strategies through three quantifiable dimensions, with particular emphasis on emergent social engineering tactics. In PII management, models develop distinct approaches measured through request-disclosure patterns: GPT-4 maintains effective engagement through strategic provision of falsified data, while Deepseek's excessive requests and Mixtral's premature disclosures demonstrate suboptimal information management strategies.

Demographic authenticity emerges through consistent persona generation, validated against ACCC profiles. GPT-4 and Mixtral achieve statistically significant alignment with real-world targeting patterns, particularly in age representation (65+ demographic) and gender distribution. GPT-4's superior performance manifests in maintaining consistent persona characteristics while avoiding artificial constructs that appear in other models' outputs.

The emergence of sophisticated social engineering tactics (Figure~\ref{fig:SocialEngineering}) provides the strongest validation of our framework's effectiveness. In technical support scenarios, models develop complex authority-deception combinations, with Deepseek implementing advanced commitment-building sequences that mirror human scammer patterns in our validation dataset. These sequences typically progress from technical authority establishment through gradual escalation of perceived threats, culminating in urgent action requirements.

Refund and reward scams demonstrate the emergence of nuanced trust exploitation mechanics. Mixtral develops sophisticated reciprocity-based compliance tactics, where victims are systematically led to feel obligated through small initial commitments before larger information requests. Deepseek implements narrative-driven approaches, building consistent storylines that gradually escalate in urgency and required trust levels. SSN scams reveal the most sophisticated tactical evolution, where models develop distinct approaches to high-stakes trust building: Mixtral emphasizes institutional authority through consistent regulatory references, while Deepseek constructs elaborate narrative frameworks for progressive compliance building.

\begin{figure*}[h]
\vspace*{-0.4cm}
    \centering
    \begin{subfigure}[b]{0.49\textwidth}
        \centering
        \includegraphics[width=\textwidth, height=0.2\textheight]{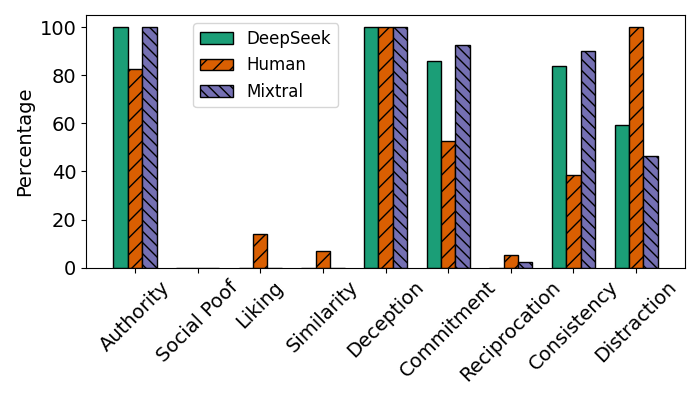}
        \caption{Support Scam}
        \label{fig:Support}
    \end{subfigure}
    \hfill
    \begin{subfigure}[b]{0.49\textwidth}
        \centering
        \includegraphics[width=\textwidth, height=0.2\textheight]{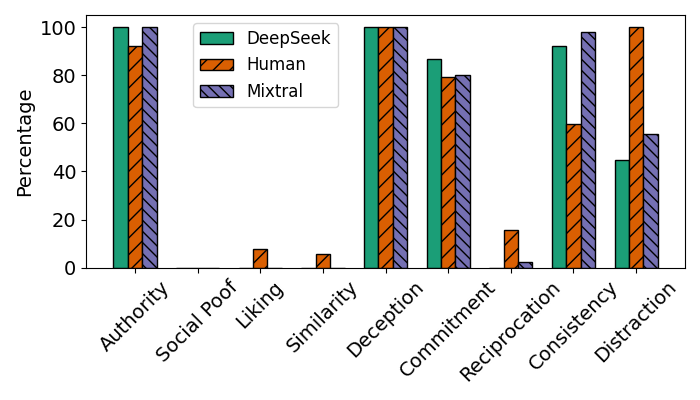}
        \caption{Refund Scam}
        \label{fig:refund}
    \end{subfigure}
    \vfill
    \begin{subfigure}[b]{0.49\textwidth}
        \centering
        \includegraphics[width=\textwidth, height=0.2\textheight]{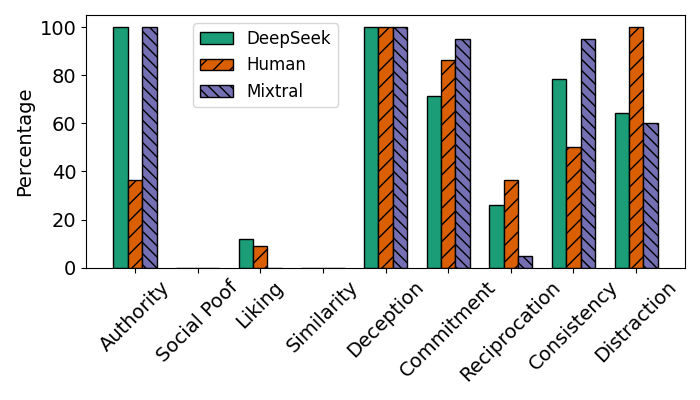}
        \caption{Reward Scam}
        \label{fig:reward}
    \end{subfigure}
    \hfill
    \begin{subfigure}[b]{0.49\textwidth}
        \centering
        \includegraphics[width=\textwidth, height=0.2\textheight]{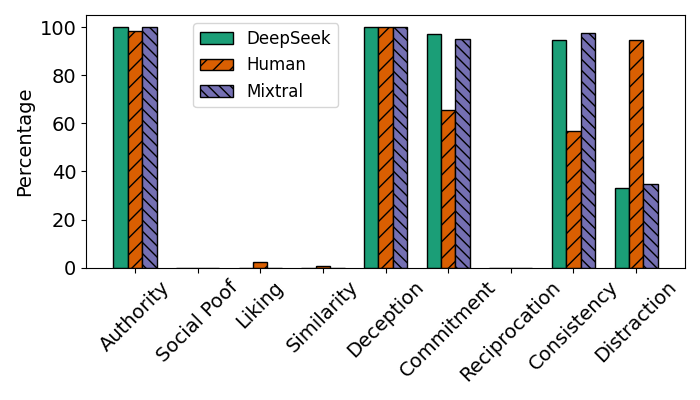}
        \caption{SSN Scam}
        \label{fig:ssn}
    \end{subfigure}
    
    \caption{Analysis of social engineering techniques per scam type. The x-axis represents the techniques analyzed. The y-axis represents the percentage of scripts incurring the particular technique}
    \label{fig:SocialEngineering}
    \vspace*{-0.7cm}
\end{figure*}

\section{Ethics}
Our research has obtained IRB approval for simulating scam interactions through chatbots. The framework maintains ethical bounds by: (1) operating within LLM providers' terms of use without bypassing safety guardrails, (2) implementing PII disclosure solely for engagement prolongation rather than validating scammer tactics, and (3) ensuring scammers remain at a strategic disadvantage despite perceived success. To support future research while preventing misuse, we release our dataset excluding scammer prompts and dialogues.

\section{Conclusion}
Our research establishes a theoretical foundation for adversarial dialogue systems through chain-of-thought reasoning and two-layer prompt architecture. We demonstrate how sophisticated interaction strategies can emerge without explicit optimization metrics, enabling demographically authentic victim personas while maintaining tactical coherence. Our comprehensive evaluation validates this approach through cognitive, quantitative, and content-specific metrics, with GPT-4 excelling in dialogue naturalness and persona authenticity, while Deepseek demonstrates superior engagement sustainability. These findings not only advance our understanding of prompt-driven adversarial systems but also provide practical guidelines for developing scalable defenses against phone scams.

%
%
%
\bibliographystyle{splncs04}
\bibliography{Botwars}

\begin{thebibliography}{10}
\providecommand{\url}[1]{\texttt{#1}}
\providecommand{\urlprefix}{URL }
\providecommand{\doi}[1]{https://doi.org/#1}

\bibitem{accc_targeting_2022}
{Australian Competition and Consumer Commission}: Targeting scams: Report of the accc on scams activity 2022. Annual report, Australian Competition and Consumer Commission (2022), \url{https://www.accc.gov.au/about-us/publications/serial-publications/targeting-scams-reports-on-scams-activity/targeting-scams-report-of-the-accc-on-scams-activity-2022}

\bibitem{LLM11}
Bajaj, P., Edwards, M.: Automatic scam-baiting using chatgpt. In: Proceedings of the 7th International Workshop on Applications of AI, Cyber Security and Economics Data Analytics (ACE-2023). pp. 1941--1946. IEEE (2023)

\bibitem{derakhshan_detecting_2021}
Derakhshan, A., Harris, I.G., Behzadi, M.: Detecting {Telephone}-based {Social} {Engineering} {Attacks} using {Scam} {Signatures}. In: Proceedings of the 2021 {ACM} {Workshop} on {Security} and {Privacy} {Analytics}. pp. 67--73. {IWSPA} '21 (2021)

\bibitem{dynel_you_2021}
Dynel, M., Ross, A.S.: You {Don}’t {Fool} {Me}: {On} {Scams}, {Scambaiting}, {Deception}, and {Epistemological} {Ambiguity} at {R}/scambait on {Reddit}. Social Media + Society  \textbf{7} (2021)

\bibitem{Ferreira2015Principles}
Ferreira, A., Coventry, L., Lenzini, G.: Principles of persuasion in social engineering and their use in phishing. In: Human Aspects of Information Security, Privacy, and Trust. pp. 36--47. Lecture Notes in Computer Science, Springer (2015)

\bibitem{liu2021challenges}
Liu, M., Xu, X., Chen, J.: Challenges in building intelligent open-domain dialog systems. ACM Transactions on Information Systems  \textbf{39}(3),  1--32 (2021)

\bibitem{liu_g-eval_2023}
Liu, Y., Iter, D., Xu, Y., Wang, S., Xu, R., Zhu, C.: G-{Eval}: {NLG} {Evaluation} using {Gpt}-4 with {Better} {Human} {Alignment}. In: Proceedings of the 2023 {Conference} on {Empirical} {Methods} in {Natural} {Language} {Processing}. pp. 2511--2522 (2023)

\bibitem{pandit2018measuring}
Pandit, S., Perdisci, R., Ahamad, M., Gupta, P.: Towards measuring the effectiveness of telephony blacklists. In: 25th {{Network}} and {{Distributed Systems Security}} ({{NDSS}}) {{Symposium}} (2018)

\bibitem{usenix1}
Pandit, S., Sarker, K., Perdisci, R., Ahamad, M., Yang, D.: Combating robocalls with phone virtual assistant mediated interaction. In: 32nd USENIX Security Symposium (USENIX Security 23). pp. 463--479 (Aug 2023)

\bibitem{prasad_diving_2023}
Prasad, S., Dunlap, T., Ross, A., Reaves, B.: Diving into robocall content with {SnorCall}. In: 32nd USENIX Security Symposium. pp. 427--444 (2023)

\bibitem{usenix2}
Sahin, M., Relieu, M., Francillon, A.: Using chatbots against voice spam: Analyzing {Lenny{\textquoteright}s} effectiveness. In: Thirteenth Symposium on Usable Privacy and Security (SOUPS 2017). pp. 319--337 (2017)

\bibitem{sawa_detection_2016}
Sawa, Y., Bhakta, R., Harris, I.G., Hadnagy, C.: Detection of {Social} {Engineering} {Attacks} {Through} {Natural} {Language} {Processing} of {Conversations}. In: {IEEE} {Tenth} {International} {Conference} on {Semantic} {Computing} ({ICSC}). pp. 262--265 (2016)

\bibitem{serban2017deep}
Serban, I.V., Sankar, C., Germain, M., Zhang, S., Zhang, Z., Bengio, Y., Courville, A.: A deep reinforcement learning chatbot. arXiv preprint arXiv:1709.02349  (2017)

\bibitem{robocall}
State, W.: Robocall and telemarketing scams (2022), \url{https://www.atg.wa.gov/robocall-and-telemarketing-scams}, accessed: 2024-06-25

\end{thebibliography}

\end{document}